\documentclass{llncs}
\usepackage{llncsdoc}
\usepackage{latexsym}
\usepackage{amsmath} %%

\usepackage[pdftex]{graphicx} %% picture
\usepackage{subfigure}
\usepackage{epstopdf}
\usepackage{pinyin}
\usepackage{CJK} %% chinese

\usepackage{multirow} %%
\usepackage{array}

\title{Look-ahead Attention for Generation in Neural Machine Translation}

\author{Long Zhou$^{\dagger}$, Jiajun Zhang$^{\dagger}$, Chengqing Zong$^{\dagger\ddagger}$ \\}

\institute{
  $^\dagger$University of Chinese Academy of Sciences \\
  National Laboratory of Pattern Recognition, CASIA  \\
  $^\ddagger$CAS Center for Excellence in Brain Science and Intelligence Technology \\
  {\tt \{long.zhou,jjzhang,cqzong\}@nlpr.ia.ac.cn} \\}

\begin{document}

%\begin{CJK*}{GBK}{song}

\maketitle

\begin{abstract}
  %%question; what we do; hao we do; great results
  The attention model has become a standard component in neural machine translation (NMT) and it guides translation process by selectively focusing on parts of the source sentence when predicting each target word. %
  However, we find that the generation of a target word does not only depend on the source sentence, but also rely heavily on the previous generated target words, especially the distant words which are difficult to model by using recurrent neural networks.
  To solve this problem, we propose in this paper a novel look-ahead attention mechanism for generation in NMT, which aims at directly capturing the dependency relationship between target words. We further design three patterns to integrate our look-ahead attention into the conventional attention model.
  Experiments on NIST Chinese-to-English and WMT English-to-German translation tasks show that our proposed look-ahead attention mechanism achieves substantial improvements over state-of-the-art baselines.

\end{abstract}

\section{Introduction}

%%question
Neural machine translation (NMT) has significantly improved the quality of machine translation in recent several years~\cite{Kalchbrenner:2013,Sutskever:2014,Bahdanau:2015,Junczys-Dowmunt:2016A},
in which the attention model increasingly plays an important role.
Unlike traditional statistical machine translation (SMT)~\cite{Koehn:2003,Chiang:2005,zhai2012tree} which contains multiple separately tuned components, NMT builds upon a single and large neural network to directly map source sentence to associated target sentence.

Typically, NMT adopts the encoder-decoder architecture which consists of two recurrent neural networks.
The encoder network models the semantics of the source sentence and transforms the source sentence into context vector representation, from which the decoder network generates the target translation word by word.
Attention mechanism has become an indispensable component in NMT, which enables the model to dynamically compose source representation for each timestep during decoding, instead of a single and static representation.
Specifically, the attention model shows which source words the model should focus on in order to predict the next target word.

%Specifically the mechanism allows the model to accord varying attention to different parts of source sentence while generating successive target words.

%The attention model plays a crucial role in NMT, as it shows which source word the model should focus on in order to predict the next target word.

\begin{figure*}[t]
    \centering
    \includegraphics[width=11cm]{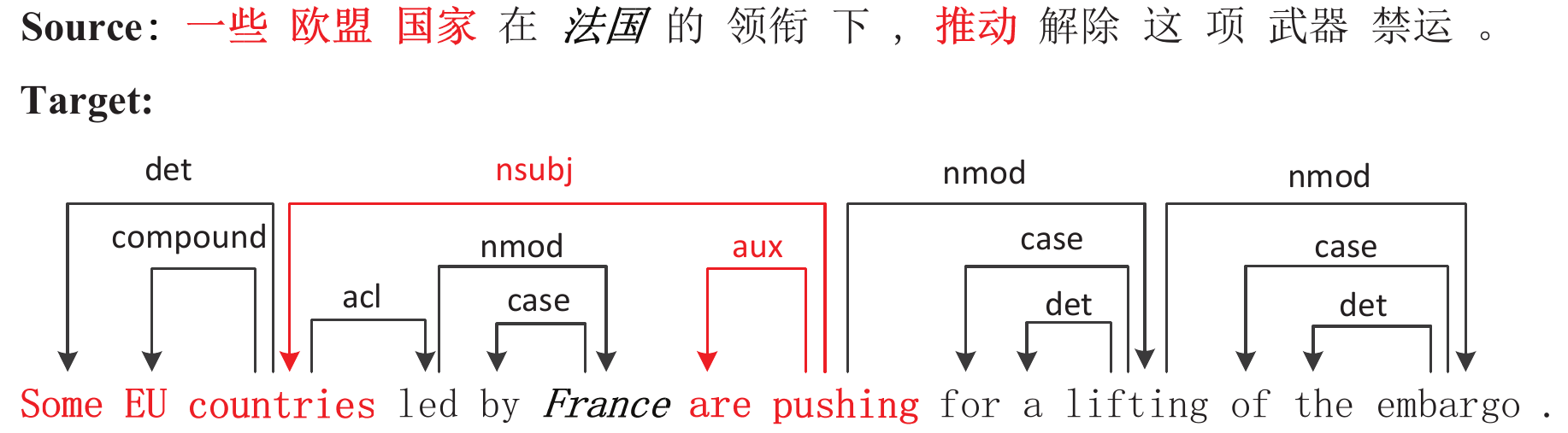}
    \caption{An example of Chinese-English translation. %\footnote{http://www.statmt.org/wmt14/translation-task.html}
    The English sentence is analyzed using Stanford online parser\protect\footnotemark[1]. %}\label{fig:0}
    Although the predicate ``\emph{are pushing}'' is close to the word ``\emph{France}'', it has a stronger dependency on the word ``\emph{countries}'' instead of ``\emph{France}''. }\label{fig:0}
    %The subject of predicate ``are pushing'' is the phrase ``some EU countries'' instead of ``France'',
    %which determines whether the predicate is singular or plural.}\label{fig:0}
\end{figure*}

%%challenge
However, previous attention models are mainly designed to predict the alignment of a target word with respect to source words, which take no account of the fact that the generation of a target word may have a stronger correlation with the previous generated target words.
Recurrent neural networks, such as gated recurrent units (GRU)~\cite{Cho:2014} and long short term memory (LSTM)~\cite{hochreiter1997long}, still suffer from long-distance dependency problems,
according to pioneering studies~\cite{Bahdanau:2015,Koehn:2017} that the performance of NMT is getting worse as source sentences get longer.
%Gated recurrent units (GRU)~\cite{Cho:2014} and long short term memory (LSTM)~\cite{hochreiter1997long} still suffer from long-distance dependency problem
Figure \ref{fig:0} illustrates an example of Chinese-English translation.
The dependency relationship of target sentence determines whether the predicate of the sentence should be singular (is) or plural (are).
While the conventional attention model does not have a specific mechanism to learn the dependency relationship between target words.
%In addition, one key factor affecting the ability to capture such dependencies is the length of the paths that forward and backward signals have to traverse in the network~\cite{hochreiter2001gradient}.

%%our work
To address this problem, we propose in this paper a novel look-ahead attention mechanism for generation in NMT, which can directly model the long-distance dependency relationship between target words.
The look-ahead attention model does not only align to source words, but also refer to the previous generated target words when generating a target word.
Furthermore, we present and investigate three patterns for the look-ahead attention, which can be integrated into any attention-based NMT.
To show the effectiveness of our look-ahead attention, we have conducted experiments on NIST Chinese-to-English translation tasks and WMT14 English-to-German translation tasks.
Experiments show that our proposed model obtains significant BLEU score improvements over strong SMT baselines and a state-of-the-art NMT baseline. %considerable

\footnotetext[1]{http://nlp.stanford.edu:8080/parser/index.jsp.}

%% for 3.1
%\begin{figure*}[t]
%    \centering
%    \includegraphics[width=11cm]{NMT.eps}
%    \caption{The architecture of neural machine translation model.}\label{fig:1}
%\end{figure*}

\section{Neural Machine Translation} \label{nmt}

\begin{figure*}[t]
    \centering
    \includegraphics[width=11cm]{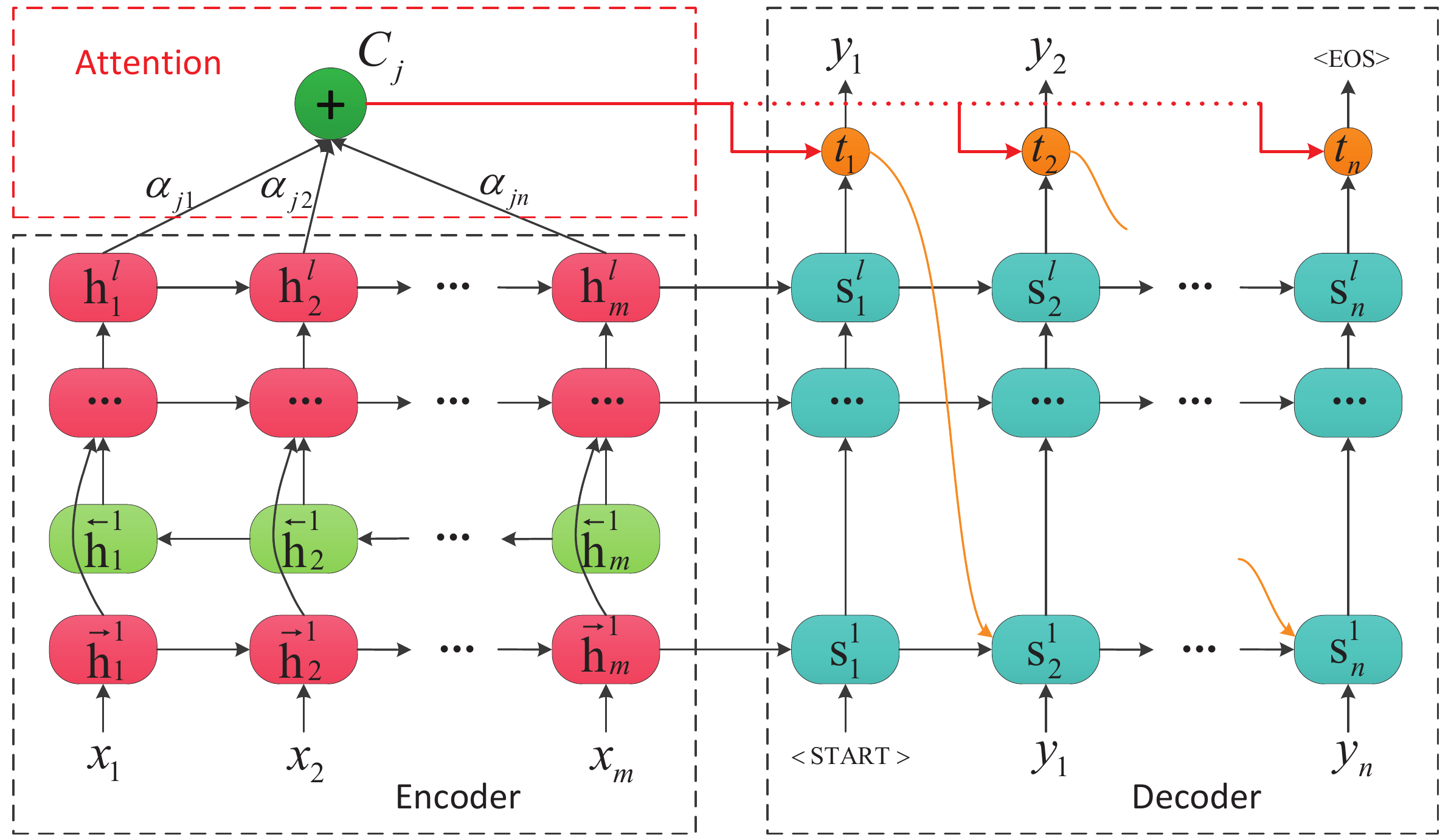}
    \caption{The architecture of neural machine translation model.}\label{fig:1}
\end{figure*}

%As a background and baseline, in this section, we briefly describe the NMT model
Our framework integrating the look-ahead attention mechanism into NMT can be applied in any conventional attention model.
Without loss of generality, we use the improved attention-based NMT proposed by Luong et al.~\cite{Luong:2015A}, which utilizes stacked LSTM layers for both encoder and decoder as illustrated in Figure \ref{fig:1}.

The NMT first encodes the source sentence $X=(x_1,x_2,...,x_m)$ into a sequence of context vector representation $C=(h_1,h_2,...,h_m)$ whose size varies with respect to the source sentence length.
Then, the NMT decodes from the context vector representation $C$ and generates target translation $Y=(y_1,y_2,...,y_n)$ one word each time by maximizing the probability of $p(y_j|y_{<j},C)$.
Next, we briefly review the encoder introducing how to obtain $C$ and the decoder addressing how to calculate $p(y_j|y_{<j},C)$.

{\bf Encoder:} The context vector representation $C=(h_1^l,h_2^l,...,h_m^l)$ are generated by the encoder using $l$ stacked LSTM layers.
Bi-directional connections are used for the bottom encoder layer, and $h_i^1$ is a concatenation of a left-to-right $\overrightarrow{h}_i^1$ and a right-to-left$\overleftarrow{h}_i^1$,
\begin{equation}
    h_i^1 = \begin{bmatrix} \overrightarrow{h}_i^1 \\ \overleftarrow{h}_i^1 \end{bmatrix} =
    \begin{bmatrix} LSTM(\overrightarrow{h}_{i-1}^1, x_i) \\ LSTM(\overleftarrow{h}_{i-1}^1,x_i) \end{bmatrix}
    %\begin{bmatrix} \overrightarrow{\o}(x_i,\overrightarrow{h}_{i-1}) \\ \overleftarrow{\o}(x_i,\overleftarrow{h}_{i+1}) \end{bmatrix}
\end{equation}

All other encoder layers are unidirectional, and $h_i^k$ is calculated as follows:
\begin{equation}
    h_i^k = LSTM(h_{i-1}^k, h_i^{k-1})
\end{equation}

{\bf Decoder:} The conditional probability $p(y_j|y_{<j},C)$ is formulated as
\begin{equation} \label{condi}
    p(y_j|Y_{<j},C) = p(y_j|Y_{<j},c_j) = softmax(W_st_j)
\end{equation}

Specifically, we employ a simple concatenation layer to produce an attentional hidden state $t_j$:
\begin{equation} \label{concat}
    t_j = tanh(W_c[s_j^l;c_j]+b) = tanh(W_c^1s_j^l+W_c^2c_j+b)
\end{equation}
where $s_j^l$ denotes the target hidden state at the top layer of a stacking LSTM.
The attention model calculates $c_j$ as the weighted sum of the source-side context vector representation, just as illustrated in the upper left corner of Figure \ref{fig:1}.
\begin{equation}
    c_j = \sum_{i=1}^{m}ATT(s_j^l,h_i^l) \cdot h_i^l =  \sum_{i=1}^{m}\alpha_{ji}h_i^l
\end{equation}
where $\alpha_{ji}$ is a normalized item calculated as follows:

\begin{equation}
    \alpha_{ji} = \frac{exp(h_i^l \cdot s_j^l)}{\sum_{i^{'}}exp(h_{i^{'}}^l \cdot s_j^l)}
%    \alpha_{ji} = \frac{a}{b}
\end{equation}

$s_j^k$ is computed by using the following formula:
\begin{equation}
    s_j^k = LSTM(s_{j-1}^k, s_j^{k-1})
\end{equation}

If $k = 1$, $s_j^1$ will be calculated by combining $t_{j-1}$ as feed input~\cite{Luong:2015A}:
\begin{equation}
    s_j^1 = LSTM(s_{j-1}^1, y_{j-1}, t_{j-1})
\end{equation}

Given the bilingual training data $D = \{(X^{(z)},Y^{(z)})\}_{z=1}^Z$, all parameters of the attention-based NMT are optimized to maximize the following conditional log-likelihood:
\begin{equation}
    L(\theta) = \frac{1}{Z} \sum_{z=1}^{Z} \sum_{j=1}^n log p(y_j^{(z)} | y_{<j}^{(z)}, X^{(z)}, \theta)
\end{equation}

\section{Model Description}

%\begin{figure*}[t]
%    \centering
%    %\caption{Different architectures of look-ahead attention.}\label{fig:2}
%    \includegraphics[width=11cm]{target-attention.eps}
%    \caption{Different architectures of look-ahead attention.
%    (a) is normal attention pattern as introduced in Eq. \ref{concat} of section \ref{nmt}.
%    (b), (c) and (d) are our three approaches which integrate look-ahead attention mechanism into attention-based NMT. }\label{fig:2}
%\end{figure*}

Learning long-distance dependencies is a key challenge in machine translation.
Although the attention model introduced above has shown its effectiveness in NMT, it takes no account of the dependency relationship between target words.
%The length of the paths that forward and backward signals have to traverse is quite key for learning
%The shorter these paths between any combination of positions in the input and output sequences, the easier it is to learn long-distance dependencies~\cite{hochreiter2001gradient}.
Hence, in order to relieve the burden of LSTM or GRU to carry on the target-side long-distance dependencies, we design a novel look-ahead attention mechanism, which directly establishes a connection between the current target word and the previous generated target words.
In this section, we will elaborate on three proposed approaches about integrating the look-ahead attention into the generation of attention-based NMT.

\subsection{Concatenation Pattern}

%{\bf Concatenation Pattern:} Figure 1 illustrates the neural system combination framework, which can take as input the source sentence and the results of MT systems.

\begin{figure*}[t]
    \centering
    %\caption{Different architectures of look-ahead attention.}\label{fig:2}
    \includegraphics[width=11cm]{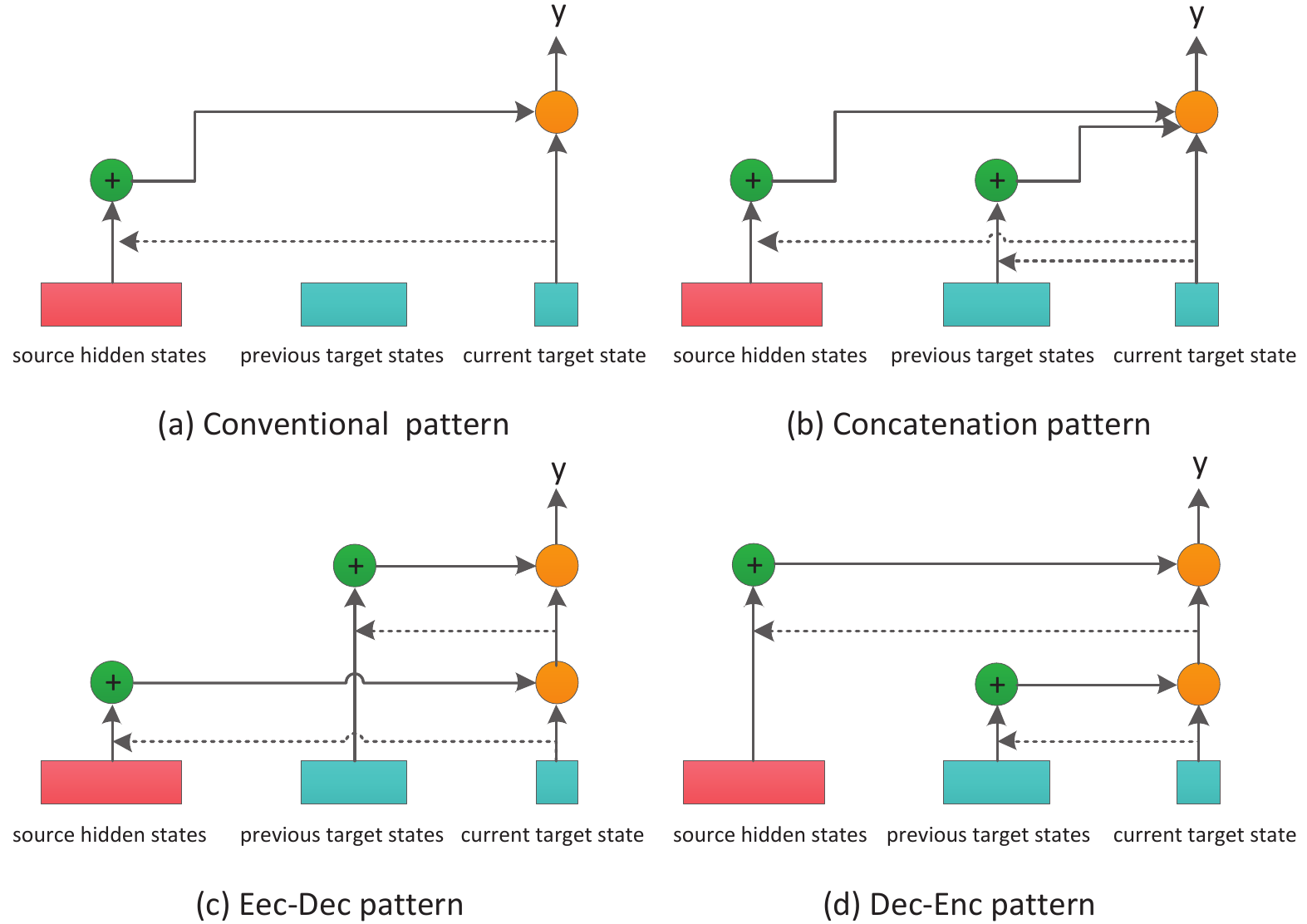}
    \caption{Different architectures of look-ahead attention.
    (a) is the conventional attention pattern as introduced in Eq. \ref{concat} of section \ref{nmt}.
    (b), (c) and (d) are our three approaches which integrate look-ahead attention mechanism into attention-based NMT. }\label{fig:2}
\end{figure*}

Figure \ref{fig:2}(b) illustrates concatenation pattern of the look-ahead attention mechanism.
We not only compute the attention between current target hidden state and source hidden states, but also calculate the attention between current target hidden state and previous target hidden states.
The look-ahead attention output at timestep $j$ is computed as:
\begin{equation} \label{datt}
    c_j^d = \sum_{i=1}^{j-1}ATT(s_j^l, s_i^l) \cdot s_i^l
\end{equation}
where $ATT(s_j^l, s_i^l)$ is a normalized item.

Specifically, given the target hidden state $s_j^l$, the source-side context vector representation $c_j$, and the target-side context vector representation $c_j^d$, we employ a concatenation layer to combine the information to produce an attentional hidden state as follows:
\begin{equation}
    t_j^{final} = tanh(W_c[s_j^l;c_j;c_j^d] + b)
\end{equation}

After getting the attentional hidden state $t_j^{final}$, we can calculate the conditional probability $p(y_j|y_{<j},C)$ as formulated in Eq. \ref{condi}.

\subsection{Enc-Dec Pattern}

Concatenation pattern is a simple method to achieve look-ahead attention, which regards source-side context vector representation and target-side context vector representation as the same importance.
Different from concatenation pattern, Enc-Dec pattern utilizes a hierarchical architecture to integrate look-ahead attention as shown in Figure \ref{fig:2}(c).

Once we get the attentional hidden state of conventional attention-based NMT, we can employ look-ahead attention mechanism to update the previous attentional hidden state.
In detail, the model first computes the attentional hidden state $t_j^e$ of conventional attention-based NMT as Eq. \ref{concat}.
Second, the model calculates the attention between the attentional hidden state $t_j^e$ and previous target hidden states:
\begin{equation}
    c_j^d = \sum_{i=1}^{j-1}ATT(t_j^e, s_i^l) \cdot s_i^l
\end{equation}

Then, the final attentional hidden state is calculated as followed:
\begin{equation}
    t_j^{final} = tanh(W_{c2}[t_j^e;c_j^d]+b_2)
\end{equation}

\subsection{Dec-Enc Pattern}

%{\bf Decoder-Encoder Pattern:} The neural system combination framework should be trained on the outputs of multiple translation systems and the gold target translations.

Dec-Enc pattern is the opposite of the Enc-Dec pattern, and it uses look-ahead attention mechanism to help the model align to source words.
Figure \ref{fig:2}(d) shows this pattern. We compute look-ahead attention output firstly as Eq. \ref{datt}, and attentional hidden state is computed by:
\begin{equation}
    t_j^d = tanh(W_{c1}[s_j^l;c_j^d]+b)
\end{equation}

Finally, we can calculate the attention between the attentional hidden state $t_j^d$ and source hidden states to get final attentional hidden state:
\begin{equation}
    t_j^{final} = tanh(W_{c2}[t_j^d;c_j^e]+b_2)
\end{equation}
\begin{equation}
    c_j^e = \sum_{i=1}^{m}ATT(t_j^d, h_i^l) \cdot h_i^l
\end{equation}
where $h_i^l$ is source-side hidden state at the top layer.

\section{Experiments}

%\subsection{Setup}
\subsection{Dataset}

We perform our experiments on the NIST Chinese-English translation tasks and WMT14 English-German translation tasks.
The evaluation metric is BLEU~\cite{Papineni:2002} as calculated by the {\small\tt multi-blue.perl} script.

For Chinese-English, our training data consists of 630K sentence pairs extracted from LDC corpus{\footnote[2]{The corpora include LDC2000T50, LDC2002T01, LDC2002E18, LDC2003E07, LDC2003E14, LDC2003T17 and LDC2004T07.}}.
We use NIST 2003(MT03) Chinese-English dataset as the validation set,
NIST 2004(MT04), NIST 2005(MT05), NIST 2006(MT06) datasets as our test sets.
Besides, 10M Xinhua portion of Gigaword corpus is used in training language model for SMT.
%For Moses, we train a 5-gram language model using target data and 11M Xinhua portion of Gigaword corpus.

For English-German, to compare with the results reported by previous work~\cite{Luong:2015A,Shen:2016,zhou2016deep}, we used the same subset of the WMT 2014 training corpus{\footnote[3]{http://www.statmt.org/wmt14/translation-task.html}} that contains 4.5M sentence pairs with 116M English words and 110M German words.
The concatenation of news-test 2012 and news-test 2013 is used as the validation set and news-test 2014 as the test set.

\subsection{Training Details}

We build the described models modified from the Zoph\_RNN{\footnote[4]{https://github.com/isi-nlp/Zoph\_RNN}} toolkit which is written in C++/CUDA and provides efficient training across multiple GPUs.
Our training procedure and hyper parameter choices are similar to those used by Luong et al.~\cite{Luong:2015A}.
In the NMT architecture as illustrated in Figure \ref{fig:1}, the encoder has three stacked LSTM layers including a bidirectional layer, followed by a global attention layer, and the decoder contains two stacked LSTM layers followed by the softmax layer.

In more details, we limit the source and target vocabularies to the most frequent 30K words for Chinese-English and 50K words for English-German.
The word embedding dimension and the size of hidden layers are all set to 1000.
Parameter optimization is performed using stochastic gradient descent(SGD), and we set learning rate to 0.1 at the beginning and halve the threshold while the perplexity go up on the development set. Each SGD is a mini-batch of 128 examples. Dropout was also applied on each layer to avoid over-fitting, and the dropout rate is set to 0.2. At test time, we employ beam search with beam size b = 12.
%For SMT, Moses-1 is an open source phrase-based translation system with default configuration and a 4-gram language model trained on the target portion of training data.

%\subsection{Training Details}

%\subsection{Translation Quality}
\subsection{Results on Chinese-English Translation}

We list the BLEU scores of our proposed model in Table \ref{c2e}.
Moses-1~\cite{koehn2007moses} is the state-of-the-art phrase-based SMT system with the default configuration and a 4-gram language model trained on the target portion of training data.
Moses-2 is the same as Moses-1 except that the language model is trained using the target data plus 10M Xinhua portion of Gigaword corpus.
%comparing with Moses~\cite{koehn2007moses} and Baseline, which is a state-of-the-art NMT as introduced in Section \ref{nmt}.
The BLEU score of our NMT baseline, which is an attention-based NMT as introduced in Section \ref{nmt}, is about 4.5 higher than the state-of-the-art SMT system Moses-2.

\begin{table}
\centering
\caption{Translation results (BLEU score) for Chinese-to-English translation.
  ``$\dagger$'': significantly better than NMT Baseline ($p<0.05$).
  ``$\ddagger$'': significantly better than NMT Baseline ($p<0.01$).} \label{c2e}
\begin{tabular}{p{3.5cm}|p{1.4cm}<{\centering}p{1.4cm}<{\centering}p{1.4cm}<{\centering}|p{1.4cm}<{\centering}}
  %\hline
  System               &      MT04   &  MT05   &   MT06   & Ave \\
  \hline
  \hline
  Moses-1                &    31.08   &   28.37   &   30.04  & 29.83   \\
  Moses-2                &    33.13   &   31.38   &   32.63  & 32.38   \\
  NMT Baseline &   38.96   &     34.95   &   36.65   &  36.85   \\
  \hline
  Concatenation pattern     &    39.43$\dagger$    &  35.40$\dagger$   &   36.93   &   37.25$\dagger$ \\
  Enc-Dec pattern      & {\bf 39.61}$\dagger$ &  {\bf 36.50}$\ddagger$ & {\bf 37.23}$\dagger$ & {\bf 37.78}$\ddagger$   \\
  Dec-Enc pattern      &    39.00    &  36.36$\ddagger$   &   37.01$\dagger$   &   37.46$\ddagger$   \\
  %\hline
\end{tabular}
%\caption{Translation results (BLEU score) for Chinese-to-English translation.} \label{c2e}
\end{table}

For the last three lines in Table \ref{c2e}, Enc-Dec pattern outperforms concatenation pattern and even Dec-Enc pattern, which shows Enc-Dec pattern is best approach to take advantage of look-ahead attention.
Moreover, our Enc-Dec pattern gets an improvement of +0.93 BLEU points over the state-of-the-art NMT baseline, which demonstrates that the look-ahead attention mechanism is effective for generation in conventional attention-based NMT.

\subsubsection{Effects of Translating Long Sentences}

\begin{figure}
    \centering
    %\caption{{\bf Length Analysis} - translation qualities(BLEU score) of our proposed model and baseline NMT as sentences become longer.}\label{fig:3}
    \includegraphics[width=8cm]{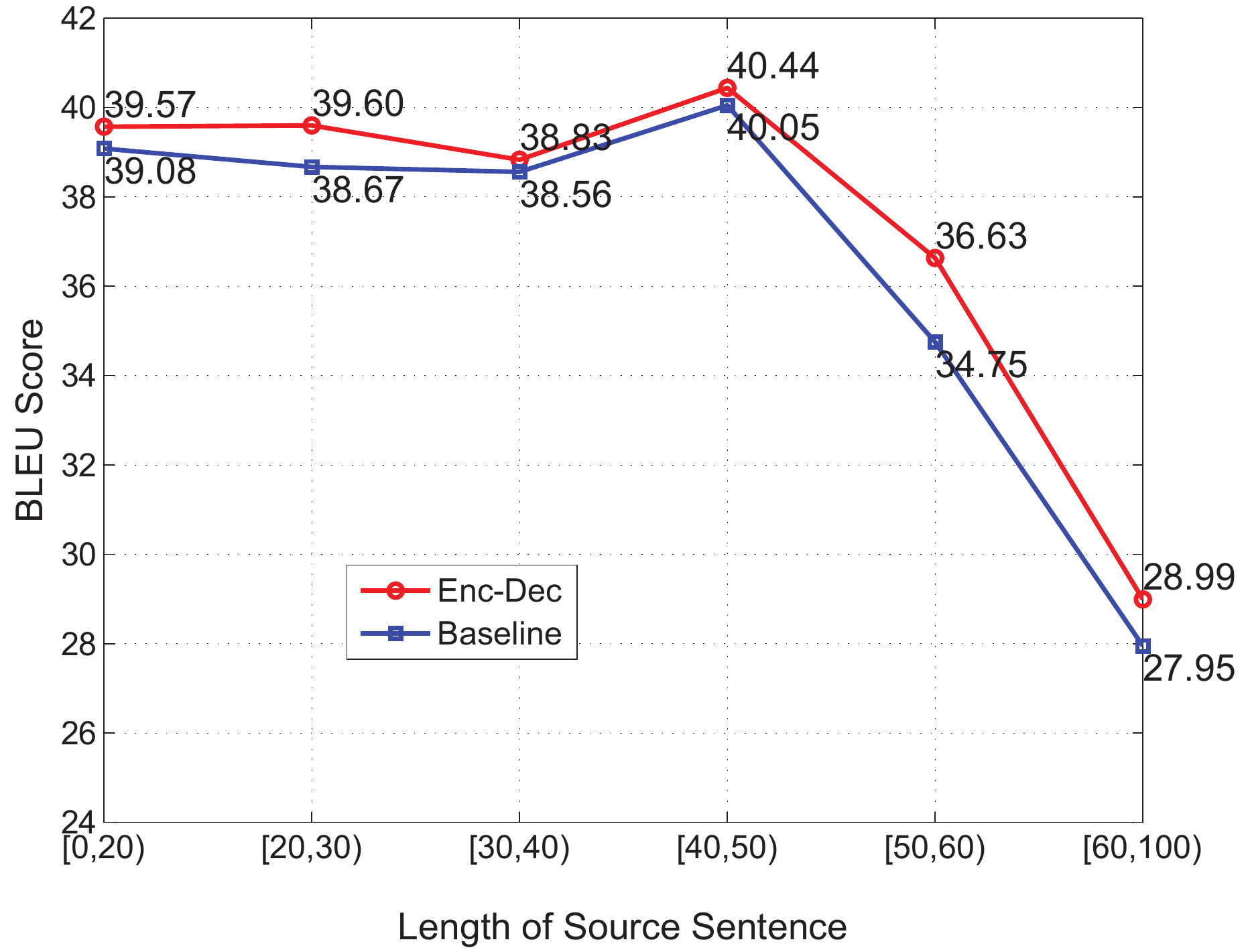}
    \caption{{\bf Length Analysis} - translation qualities(BLEU score) of our proposed model and the NMT baseline as sentences become longer.}\label{fig:3}
\end{figure}

A well-known flaw of NMT model is the inability to properly translate long sentences.
One of the goals that we integrate the look-ahead attention into the generation of NMT decoder is boosting the performance in translating long sentence. We follow Bahdanau et al.~\cite{Bahdanau:2015} to group sentences of similar lengths together and compute a BLEU score per group, as demonstrated in Figure \ref{fig:3}.

Although the performance of both the NMT baseline and our proposed model drops rapidly when the length of source sentence increases,
%NMT baseline has lower translation quality on very long sentences where words of sentence are more than 50.
our Enc-Dec model is more effective than the NMT Baseline in handling long sentences. %, as demonstrated in Figure \ref{fig:3}.
Specifically, our proposed model gets an improvement of 1.88 BLEU points over the baseline from 50 to 60 words in source language.
Furthermore, when the length of input sentence is greater than 60, our model still outperforms the baseline by 1.04 BLEU points.
Experiments show that the look-ahead attention can relieve the burden of LSTM to carry on the target-side long-distance dependencies.

\subsubsection{Target Alignment of Look-ahead Attention}

The conventional attention models always refer to some source words when generating a target word.
We propose a look-ahead attention for generation in NMT, which also focuses on previous generated words in order to predict the next target word.

\begin{figure}
    \centering
    %\caption{{\bf Length Analysis} - translation qualities(BLEU score) of our proposed model and baseline NMT as sentences become longer.}\label{fig:3}
    \includegraphics[width=12cm]{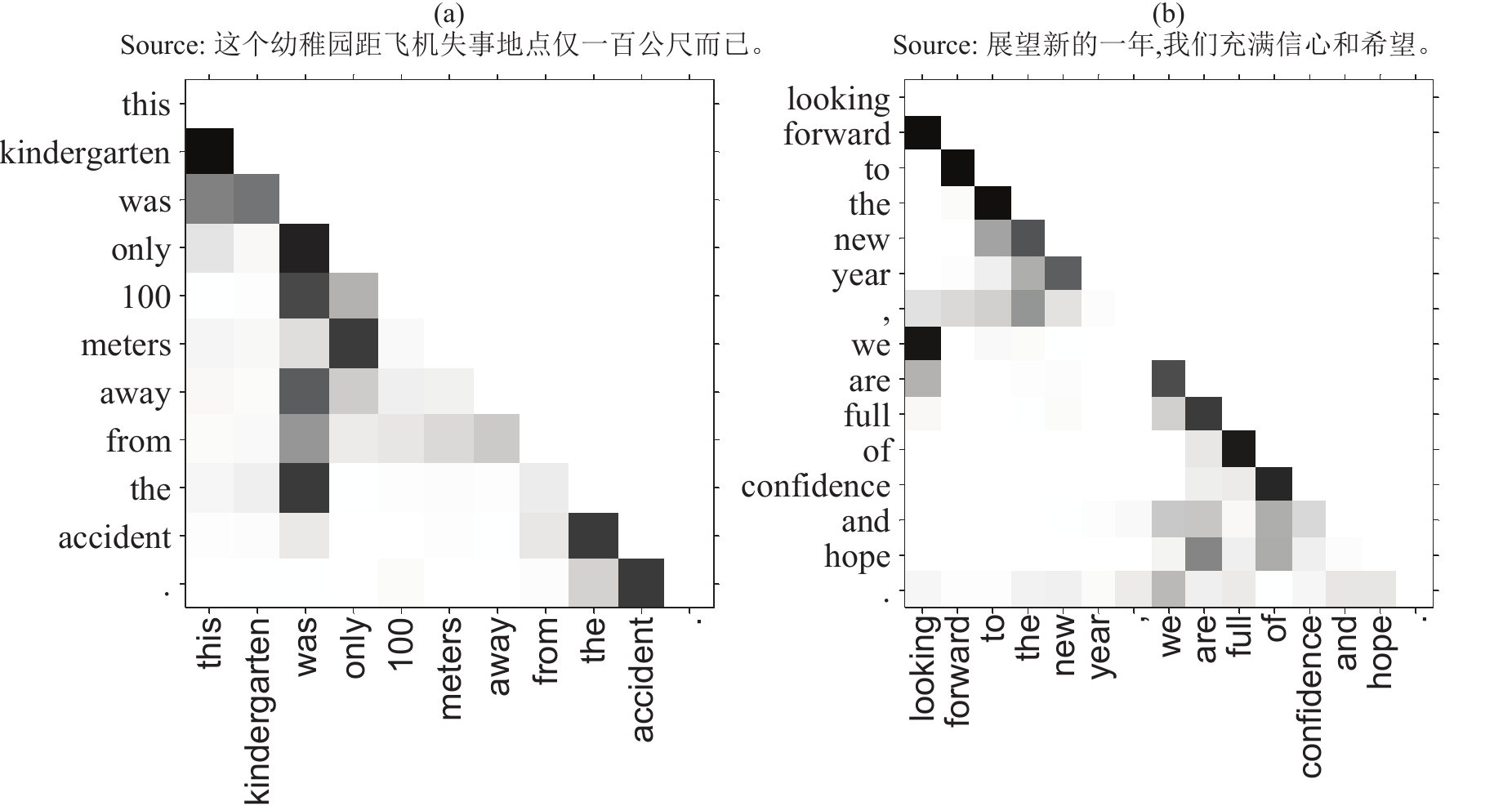}
    \caption{Target Alignment of Look-ahead Attention.}\label{fig:4}
\end{figure}

%\begin{figure}
%    \centering
%    %\caption{Target Alignment of Look-ahead Attention.}\label{fig:4}
%    \begin{minipage}{6cm}
%        %\includegraphics[width=7cm]{420.png}
%        \includegraphics[width=7cm]{untitled.eps}
%    \end{minipage}
%    \begin{minipage}{6cm}
%        \includegraphics[width=7cm]{856.png}
%    \end{minipage}
%    \caption{Target Alignment of Look-ahead Attention.}\label{fig:4}
%\end{figure}

We provide two real translation examples to show the target alignment of look-ahead attention in Figure \ref{fig:4}.
%Figure \ref{fig:4} shows two examples of target alignment in look-ahead attention.
The first line is blank because it does not have look-ahead attention when generating the first word.
Every line represents the weight distribution for previous generated words when predicting current target word.
%More specifically, we find some interesting phenomenons. First, most target words mainly focuses on the word in front of the target word. We speculate that the last generated word has most information.
%More specifically, we find that most target words mainly focuses on the word in front of the target word. We speculate that the last generated word has most information.
More specifically, we find some interesting phenomena.
First, target words often refer to verb or predicate which has been generated previously, such as the word ``was'' in Figure \ref{fig:4}(a).
%For example, the phrase ``\emph{only 100 meters away from}'' is focus on the word ``\emph{was}'' in Figure \ref{fig:4}(a).

Second, the heat map shows that the word ``we'' and the word ``looking'' have a stronger correlation when translating the Chinese sentence as demonstrated in Figure \ref{fig:4}(b).
Intuitively, the look-ahead attention mechanism establishes a bridge to capture the dependency relationship between target words.
%as shown in  Figure \ref{fig:4}(b), the word ``\emph{we}'' is the informal subject of first clauses.
%The head map shows that they have greater correlation in term of dependency relationship.
Third, most target words mainly focus on the word immediately before the current target word, which may be due to the fact that the last generated word contains more information in recurrent neural networks.
%We speculate that the last generated word has most information.
We can control the influence of the look-ahead attention like Tu et al.~\cite{tu2016context} to improve translation quality and instead we remain it as our future work.
%Intuitively, we think look-ahead attention are specific in target language, and it can extract dependence relationship between target word sometimes.

%Third, look-ahead attention model can learn anaphora relation to some extent, especially zero anaphora.
%The right part of Figure \ref{fig:4} illustrates one example of anaphora relation, and the word \emph{we} is the informal subject of first clauses.

\subsection{Results on English-German Translation}

We evaluate our model on the WMT14 translation tasks for English to German, whose results are presented in Table \ref{e2d}. We find that our proposed look-ahead attention NMT model also obtains significant accuracy improvements on large-scale English-German translation.

\begin{table}
\centering
\caption{Translation results (BLEU score) for English-to-German translation.
  ``$\ddagger$'': significantly better than Baseline($p<0.01$).} \label{e2d}
\begin{tabular}{p{2.5cm}|p{5.7cm}<{\centering}|p{1.2cm}<{\centering}|p{1.2cm}<{\centering}}
  %\hline
  System                 &          Architecture     & Voc.   & BLEU \\
  \hline
  \hline
  \multicolumn{4}{c}{Existing systems} \\
  \hline
  Loung et al.~\cite{Luong:2015A}     &  LSTM with 4 layers+dropout+local att. & 50K  &   19.00   \\
  Shen et al.~\cite{Shen:2016}    &   Gated RNN with search + MRT            & 50K &   18.02   \\
  Zhou et al.~\cite{zhou2016deep}    &   LSTM with 16 layers + F-F connections  & 160K &   20.60    \\
  \hline
  \multicolumn{4}{c}{Our NMT systems} \\
  \hline
  This work           &   Baseline                              & 50K &   19.84   \\
  This work           &   Enc-Dec pattern               & 50K &   20.36$\ddagger$   \\
  %\hline
\end{tabular}

\end{table}

In addition, we compare our NMT systems with various other systems including Zhou et al.~\cite{zhou2016deep} which use a much deeper neural network.
Luong et al.~\cite{Luong:2015A} achieves BLEU score of 19.00 with 4 layers deep Encoder-Decoder model.
Shen et al.~\cite{Shen:2016} obtained the BLEU score of 18.02 with MRT techniques.
For this work, our Enc-Dec look-ahead attention NMT model with two layers achieves 20.36 BLEU scores, which is on par with Zhou et al.~\cite{zhou2016deep} in term of BLEU.
Note that Zhou et al.~\cite{zhou2016deep} employ a much larger depth as well as vocabulary size to obtain their best results.

%The recently proposed neural machine translation has drawn more and more attention. Most of the existing approaches and models mainly focus on
%designing better attention models~\cite{Luong:2015A,Mi:2016A,Mi:2016B,Tu:2016,Meng:2016},
%better strategies for handling rare and unknown..................... words

\section{Related Work}

The recently proposed neural machine translation has drawn more and more attention. Most of the existing approaches and models mainly focus on
designing better attention models~\cite{Luong:2015A,Mi:2016A,Mi:2016B,Tu:2016,Meng:2016},
better strategies for handling rare and unknown words~\cite{Luong:2015B,Li:2016,Sennrich:2016A},
exploiting large-scale monolingual data~\cite{Cheng:2016,Sennrich:2016B,Zhang:2016B},
and integrating SMT techniques~\cite{Shen:2016,He:2016,Zhou:2017,wang2017neural}.

Our goal in this work is to design a smart attention mechanism to model the dependency relationship between target words.
Tu et al.~\cite{Tu:2016} and Mi et al.~\cite{Mi:2016A}  proposed to extend attention models with a coverage vector in order to attack the problem of repeating and dropping translations.
Cohn et al.~\cite{Cohn:2016} augmented the attention model with well-known features in traditional SMT. % ,  including positional bias, Markov conditioning, fertility and agreement over translation directions.
%To address the lack of effective control on the influence from source and target contexts, Tu et al.~\cite{tu2016context} proposed to use context gates to dynamically control the ratios at which source and target contexts contribute to the generation of target words.
Unlike previous works that attention models are mainly designed to predict the alignment of a target word with respect to source words, we focus on establishing a direct bridge to capture the long-distance dependency relationship between target words.
In addition, Wu et al.~\cite{wusequence:2016} lately proposed a sequence-to-dependency NMT method, in which the target word sequence and its corresponding dependency structure are jointly constructed and modeled. However, the target dependency tree references are needed for training in this model and our proposed model does not need extra resources.

Very Recently, Vaswani et al.~\cite{Vaswani:2017} proposed a new simple network architecture, Transformer, based solely on attention mechanisms with multi-headed self attention.
%The transduction model compute representations of its input and output without using recurrence and convolutions.
Besides, Lin et al.~\cite{lin2017structured} presented a self-attention mechanism which extracts different aspects of the sentence into multiple vector representations.
And the self-attention model has been used successfully in some tasks including abstractive summarization and reading comprehension\cite{paulus2017deep,cheng2016long}.
Here, in order to alleviate the burden of LSTM to carry on the target-side long-distance dependencies of NMT, we propose to integrate the look-ahead attention mechanism into the conventional attention-based NMT which is used in conjunction with a recurrent network.

\section{Conclusion}

In this work, we propose a novel look-ahead attention mechanism for generation in NMT, which aims at directly capturing the long-distance dependency relationship between target words.
The look-ahead attention model not only aligns to source words, but also refers to the previous generated words when generating the next target word.
Furthermore, we present and investigate three patterns to integrate our proposed look-ahead attention into the conventional attention model.
Experiments on Chinese-to-English and English-to-German translation tasks show that our proposed model obtains significant BLEU score gains over strong SMT baselines and a state-of-the-art NMT baseline.

\section*{Acknowledgments}

%We thank the anonymous reviewers for their valuable comments.
The research work has been funded by the Natural Science Foundation of China under Grant No. 61673380, No. 61402478 and No. 61403379.
% and it is also supported by the Strategic Priority Research Program of the CAS under Grant No. XDB02070007.

%\bibliographystyle{splncs}

%\begin{thebibliography}{}
%
%\bibitem[1982]{clar:eke}
%Clarke, F., Long Zhou.:
%Neural System Combinaiton for Machine Translation
%ACL
%2017(HaHa)
%
%\bibitem[2017]{clar:Zhang}
%Zhang ,J. J., Long Zhou.:
%Exploiting Source-side Monolingual Data in Neural Machine Translation
%EMNLP
%2017 ACL
%
%\end{thebibliography}

\bibliography{nlpcc}

\begin{thebibliography}{10}
\providecommand{\url}[1]{\texttt{#1}}
\providecommand{\urlprefix}{URL }

\bibitem{Bahdanau:2015}
Bahdanau, D., Cho, K., Bengio, Y.: Neural machine translation by jointly
  learning to align and translate. In Proceedings of ICLR 2015 (2015)

\bibitem{cheng2016long}
Cheng, J., Dong, L., Lapata, M.: Long short-term memory-networks for machine
  reading. arXiv preprint arXiv:1601.06733 (2016)

\bibitem{Cheng:2016}
Cheng, Y., Xu, W., He, Z., He, W., Wu, H., Sun, M., Liu, Y.: Semi-supervised
  learning for neural machine translation. In Proceedings of ACL 2016 (2016)

\bibitem{Chiang:2005}
Chiang, D.: A hierarchical phrase-based model for statistical machine
  translation. In Proceedings of ACL 2005 (2005)

\bibitem{Cho:2014}
Cho, K., van Merrienboer, B., Gulcehre, C., Bahdanau, D., Bougares, F.,
  Schwenk, H., Bengio, Y.: Learning phrase representations using RNN
  encoder¨Cdecoder for statistical machine translation. In Proceedings of EMNLP
  2014 (2014)

\bibitem{Cohn:2016}
Cohn, T., Hoang, C.D.V., Vymolova, E., Yao, K., Dyer, C., Haffari, G.:
  Incorporating structural alignment biases into an attentional neural
  translation model. arXiv preprint arXiv:1601.01085 (2016)

\bibitem{He:2016}
He, W., He, Z., Wu, H., Wang, H.: Improved neural machine translation with SMT
  features. In Proceedings of AAAI 2016 (2016)

\bibitem{hochreiter1997long}
Hochreiter, S., Schmidhuber, J.: Long short-term memory, vol.~9. MIT Press
  (1997)

\bibitem{Junczys-Dowmunt:2016A}
Junczys-Dowmunt, M., Dwojak, T., Hoang, H.: Is neural machine translation ready
  for deployment? A case study on 30 translation directions. In Proceedings of
  IWSLT 2016 (2016)

\bibitem{Kalchbrenner:2013}
Kalchbrenner, N., Blunsom, P.: Recurrent continuous translation models. In
  Proceedings of EMNLP 2013 (2013)

\bibitem{koehn2007moses}
Koehn, P., Hoang, H., Birch, A., Callison-Burch, C., Federico, M., Bertoldi,
  N., Cowan, B., Shen, W., Moran, C., Zens, R., et~al.: Moses: Open source
  toolkit for statistical machine translation. Association for Computational
  Linguistics (2007)

\bibitem{Koehn:2017}
Koehn, P., Knowles, R.: Six challanges for neural machine translation. arXiv
  preprint arXiv:1706.03872 (2017)

\bibitem{Koehn:2003}
Koehn, P., Och, F.J., Marcu, D.: Statistical phrase-based translation. In
  Proceedings of ACL-NAACL 2013 (2003)

\bibitem{Li:2016}
Li, X., Zhang, J., Zong, C.: Towards zero unknown word in neural machine
  translation. In Proceedings of IJCAI 2016 (2016)

\bibitem{lin2017structured}
Lin, Z., Feng, M., Santos, C.N.d., Yu, M., Xiang, B., Zhou, B., Bengio, Y.: A
  structured self-attentive sentence embedding. arXiv preprint arXiv:1703.03130
  (2017)

\bibitem{Luong:2015A}
Luong, M.T., Pham, H., Manning, C.D.: Effective approaches to attention-based
  neural machine translation. In Proceedings of EMNLP 2015 (2015)

\bibitem{Luong:2015B}
Luong, M.T., Sutskever, I., Le, Q.V., Vinyals, O., Zaremba, W.: Addressing the
  rare word problem in neural machine translation. In Proceedings of ACL 2015
  (2015)

\bibitem{Meng:2016}
Meng, F., Lu, Z., Li, H., Liu, Q.: Interactive attention for neural machine
  translation. In Proceedings of COLING 2016 (2016)

\bibitem{Mi:2016A}
Mi, H., Sankaran, B., Wang, Z., Ge, N., Ittycheriah, A.: A coverage embedding
  model for neural machine translation. In Proceedings of EMNLP 2016 (2016)

\bibitem{Mi:2016B}
Mi, H., Wang, Z., Ge, N., Ittycheriah, A.: Supervised attentions for neural
  machine translation. In Proceedings of EMNLP 2016 (2016)

\bibitem{Papineni:2002}
Papineni, K., Roukos, S., Ward, T., Zhu, W.: Bleu: a methof for automatic
  evaluation of machine translation. In Proceedings of ACL 2002 (2002)

\bibitem{paulus2017deep}
Paulus, R., Xiong, C., Socher, R.: A deep reinforced model for abstractive
  summarization. arXiv preprint arXiv:1705.04304 (2017)

\bibitem{Sennrich:2016B}
Sennrich, R., Haddow, B., Birch, A.: Improving neural machine translation
  models with monolingual data. In Proceedings of ACL 2016 (2016)

\bibitem{Sennrich:2016A}
Sennrich, R., Haddow, B., Birch, A.: Neural machine translation of rare words
  with subword units. In Proceedings of ACL 2016 (2016)

\bibitem{Shen:2016}
Shen, S., Cheng, Y., He, Z., He, W., Wu, H., Sun, M., Liu, Y.: Minimum risk
  training for neural machine translation. In Proceedings of ACL 2016 (2016)

\bibitem{Sutskever:2014}
Sutskever, I., Vinyals, O., Le, Q.V.: Sequence to sequence learning with neural
  networks. In Proceedings of NIPS 2014 (2014)

\bibitem{tu2016context}
Tu, Z., Liu, Y., Lu, Z., Liu, X., Li, H.: Context gates for neural machine
  translation. arXiv preprint arXiv:1608.06043 (2016)

\bibitem{Tu:2016}
Tu, Z., Lu, Z., Liu, Y., Liu, X., Li, H.: Modeling coverage for neural machine
  translation. In Proceedings of ACL 2016 (2016)

\bibitem{Vaswani:2017}
Vawani, A., Shazeer, N., Parmar, N., Uszkoreit, J., Jones, L., N.Gomez, A.,
  Kaiser, L., Polosukhin, I.: Attention is all you need. arXiv preprint
  arXiv:1706.03762 (2016)

\bibitem{wang2017neural}
Wang, X., Lu, Z., Tu, Z., Li, H., Xiong, D., Zhang, M.: Neural machine
  translation advised by statistical machine translation. In Proceedings of
  AAAI 2017 (2017)

\bibitem{wusequence:2016}
Wu, S., Zhang, D., Yang, N., Li, M., Zhou, M.: Sequence-to-dependency neural
  machine translation. In Proceedings of ACL 2017 (2017)

\bibitem{zhai2012tree}
Zhai, F., Zhang, J., Zhou, Y., Zong, C., et~al.: Tree-based translation without
  using parse trees. In Proceedings of COLING 2012 (2012)

\bibitem{Zhang:2016B}
Zhang, J., Zong, C.: Exploiting source-side monolingual data in neural machine
  translation. In Proceedings of EMNLP 2016 (2016)

\bibitem{zhou2016deep}
Zhou, J., Cao, Y., Wang, X., Li, P., Xu, W.: Deep recurrent models with
  fast-forward connections for neural machine translation. arXiv preprint
  arXiv:1606.04199 (2016)

\bibitem{Zhou:2017}
Zhou, L., Hu, W., Zhang, J., Zong, C.: Neural system combination for machine
  translation. In Proceedings of ACL 2017 (2017)

\end{thebibliography}
\bibliographystyle{splncs03}

%\end{CJK*}
\end{document}